\documentclass[lettersize,journal]{IEEEtran}
\usepackage{amsmath,amsfonts}
\usepackage{algorithmic}
\usepackage{algorithm}
\usepackage{array}
\usepackage[caption=false,font=normalsize,labelfont=sf,textfont=sf]{subfig}
\usepackage{textcomp}
\usepackage{stfloats}
\usepackage{url}
\usepackage{verbatim}
\usepackage{graphicx}
\usepackage{cite}
\usepackage{multirow}
\usepackage{booktabs,tabularx}
\usepackage{amssymb}
\usepackage[switch]{lineno}
\usepackage{newfloat}
\usepackage{listings}
\usepackage[pagebackref,breaklinks,colorlinks]{hyperref}
\hypersetup{citecolor=[RGB]{119,185,0}}

\newcommand{\green}[1]{\textcolor[RGB]{96,177,87}{#1}}
\newcommand{\red}[1]{\textcolor[RGB]{227,96,87}{#1}}

\begin{document}

\title{FAST: Faster Arbitrarily-Shaped Text Detector with Minimalist Kernel Representation}

\author{Zhe Chen, Jiahao Wang, Wenhai Wang, Guo Chen, Enze Xie, Ping Luo, Tong Lu
\thanks{
Zhe Chen, Jiahao Wang, Guo Chen, and Tong Lu are with the Department of Computer Science and Technology, Nanjing University, Jiangsu, P. R. China (E-mail: chenzhe98@smail.nju.edu.cn; wangjh@smail.nju.edu.cn; chenguo1177@gmail.com; lutong@nju.edu.cn).}
\thanks{
Wenhai Wang is with the Shanghai AI Laboratory, Shanghai, P. R. China (E-mail: wangwenhai@pjlab.org.cn).}
\thanks{
Enze Xie and Ping Luo are with the Department of Electronic Engineer, The Chinese University of Hong Kong, Hong Kong, P. R. China (E-mail: Johnny\_ez@163.com; pluo@ie.cuhk.edu.hk).}
}

\markboth{Journal of \LaTeX\ Class Files,~Vol.~14, No.~8, August~2021}%
{Shell \MakeLowercase{\textit{et al.}}: A Sample Article Using IEEEtran.cls for IEEE Journals}


\def\ie{\textit{i.e.}}
\def\eg{\textit{e.g.}}
\def\etc{etc}
\def\etal{\textit{et al.}}

\maketitle

\begin{abstract}
We propose an accurate and efficient scene text detection framework, termed FAST (\ie, faster arbitrarily-shaped text detector).
Different from recent advanced text detectors that used complicated post-processing and hand-crafted network architectures, resulting in low inference speed, FAST has two new designs.
(1) We design a minimalist kernel representation (only has 1-channel output) to model text with arbitrary shape, as well as a GPU-parallel post-processing to efficiently assemble text lines with a negligible time overhead.
(2) We search the network architecture 
tailored for text detection, leading to more powerful features than most networks that are searched for image classification.
Benefiting from these two designs, FAST achieves an excellent trade-off between accuracy and efficiency on several challenging datasets, including Total Text, CTW1500, ICDAR 2015, and MSRA-TD500.
For example, FAST-T yields 81.6\% F-measure at 152 FPS on Total-Text, outperforming the previous fastest method by 1.7 points and 70 FPS in terms of accuracy and speed.
With TensorRT optimization, the inference speed can be further accelerated to over 600 FPS.
Code and models will be released at \url{https://github.com/czczup/FAST}.
\end{abstract}


\section{Introduction}

Scene text detection is a fundamental task in computer vision with wide practical applications, such as image understanding, instant translation, and autonomous driving.
With the remarkable progress of deep learning, a considerable amount of methods \cite{liao2020real,long2018textsnake,wang2019shape,wang2019efficient,ye2020textfusenet,zhu2021fourier} have been proposed to detect text with arbitrary shape, and the performance on public datasets is constantly being refreshed.
However, we argue that the above methods still have room to improve due to two main sub-optimal designs:
(1) low-efficient post-processing and 
(2) hand-crafted network architecture.

\begin{figure}[tbp]
\renewcommand{\arraystretch}{1}
    \centering
    \small
    \includegraphics[width=0.98\linewidth]{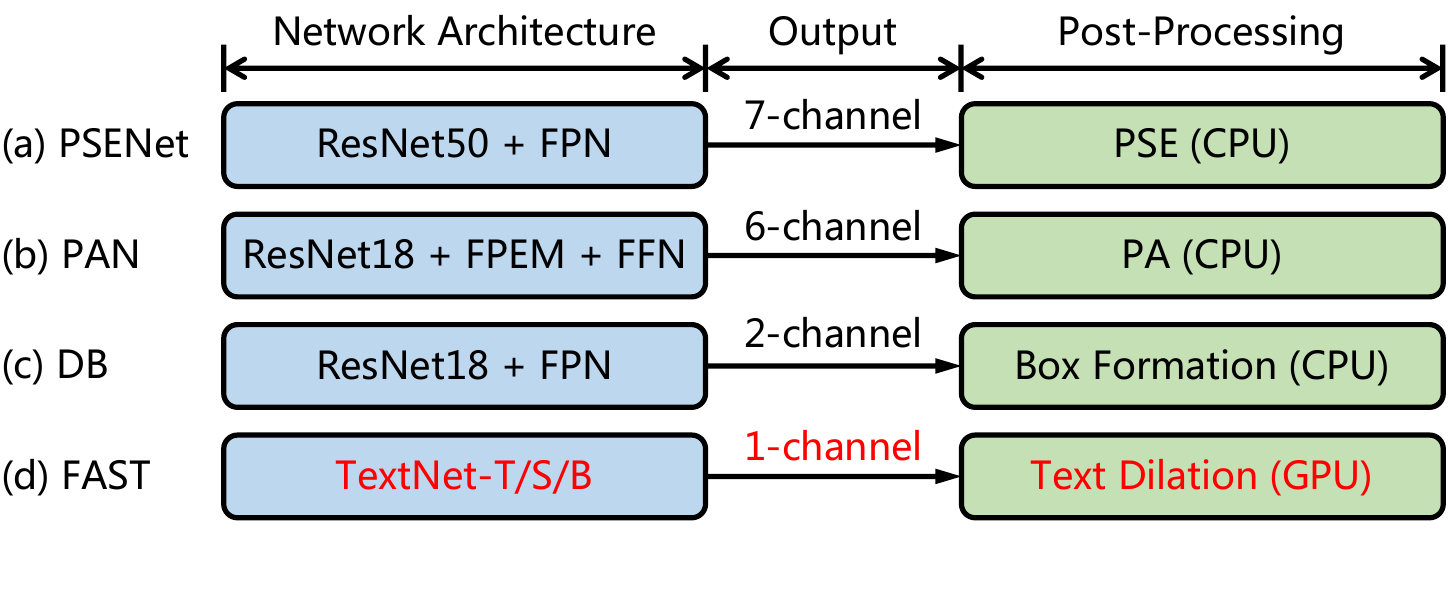}
    \setlength{\tabcolsep}{2.5mm}{
        \begin{tabular}{lcccc}
        \toprule
        \multirow{2}{*}{Method} & \multirow{2}{*}{F(\%)} & \multicolumn{2}{c}{Time Cost (ms)} & \multirow{2}{*}{FPS} \\
         &  & Network & Post-Proc. & \\
        \midrule  
        PSENet-1s~\cite{wang2019shape} & 80.9 & 118.0 & 145.0 & 3.9  \\
        PAN-512~\cite{wang2019efficient} & 84.3 & 13.7 & 3.5 & 57.1 \\
        DB-R18~\cite{liao2020real} & 82.8 & 15.3 & 4.7 & 50.0  \\
        FAST-S-512 (ours) & \textbf{84.9} & \textbf{7.5} & \textbf{1.1} & \textbf{115.5}  \\
        \bottomrule
        \end{tabular}}
    \vspace{0.7em}
    \caption{
    Overall pipelines of representative arbitrarily-shaped text detectors.
    ``Post-Proc." is short for post-processing.
    Our FAST achieves significantly faster inference speed than previous methods~\cite{wang2019shape, wang2019efficient, liao2020real}, benefiting from (1) the minimalist kernel representation (MKR) with a GPU-parallel post-processing method---text deliation, and (2) the efficient TextNet architecture specifically searched for text detection.
    }
    \label{fig:pipeline}
\end{figure}

First, the post-processing of previous works usually takes about 30\% of the whole inference time \cite{liao2020real,wang2019shape,wang2019efficient}.
Moreover, these post-processing approaches are designed to run on the CPU (see Fig.~\ref{fig:pipeline}), which are difficult to parallel with GPU resources, resulting in relatively low efficiency.
In general, the post-processing is closely related to the text representation method~\cite{liao2020real,liu2020abcnet,wang2019shape,wang2019efficient,long2018textsnake}, which determines whether it can be optimized to achieve GPU parallelism.
Therefore, it is important to develop a GPU-friendly representation method with a parallelable post-processing for the real-time text detector.

\begin{figure}[!t]
    \centering
    \includegraphics[width=1\linewidth]{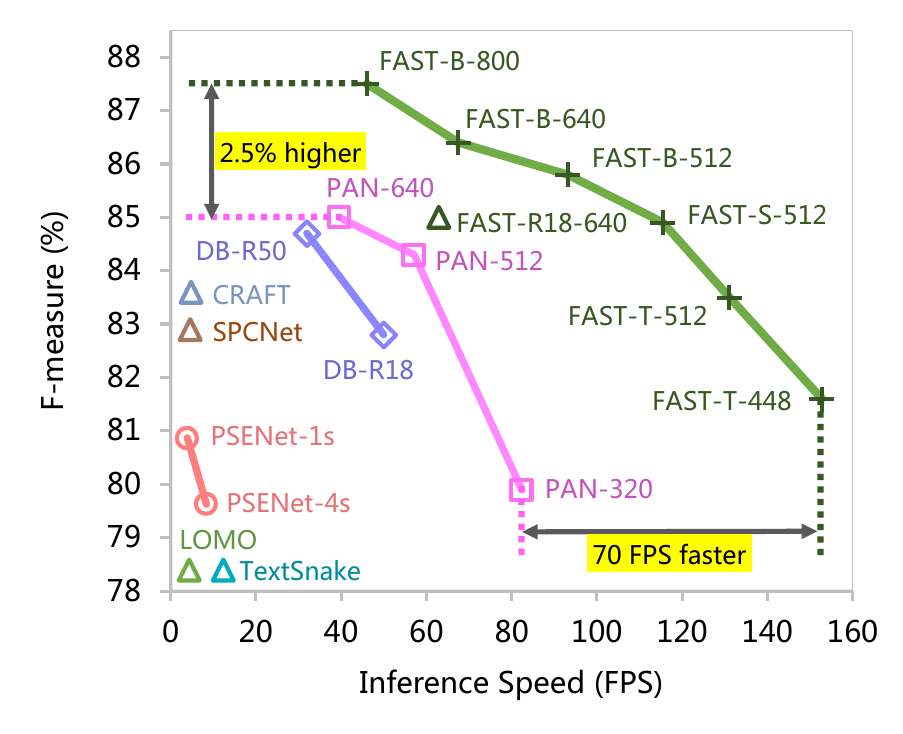}
    \caption{
    Text detection F-measure and inference speed of different text detectors on 
    Total-Text~\cite{ch2017total} dataset.
    Our FAST models, namely FAST-T/S/B according to the model size, enjoy faster inference speed and better accuracy than state-of-the-art counterparts, such as PAN~\cite{wang2019efficient} and DB~\cite{liao2020real}.
    }
    \label{fig:f_vs_speed}
\end{figure}

Second, most existing text detectors adopt a heavy hand-crafted backbone (\eg, ResNet50~\cite{he2016deep}) to achieve excellent performance, but at the expense of inference speed to some extent. 
For high efficiency, some methods \cite{liao2020real,wang2019efficient} developed text detectors based on ResNet18 \cite{he2016deep}, but the backbone is originally designed for image classification and may not be the best choice for text detection.
Although many auto-searched lightweight networks \cite{cai2019once,cai2018proxylessnas,howard2019searching,tan2019efficientnet} have been presented, they only focus on image classification or general object detection, and the application to text detection is rarely considered.
Consequently, how to design an efficient and powerful network specific to text detection, is a topic worth exploring.

In this work, we propose an efficient and powerful text detection framework, termed FAST (\textbf{F}aster \textbf{A}rbitrary-\textbf{S}haped \textbf{T}ext detector).
As illustrated in Fig.~\ref{fig:pipeline}, FAST contains the following two main improvements to achieve high efficiency:
(1) We propose a minimalist kernel representation (MKR) that formulates a text line as an eroded text region surrounded by peripheral pixels.
Compared to existing kernel representations~\cite{liao2020real,wang2019shape,wang2019efficient}, our MKR not only benefits the network to predict a 1-channel output, but also enjoys a GPU-parallel post-processing---text dilation.
(2) We carefully design a NAS search space and reward function for the text detection task.
The searched efficient backbones are named TextNet, which can provide more powerful features for text detection than the network searched on image classification (\eg, MobileNetV3~\cite{howard2019searching}).
Combining the advantages of these designs, our method achieves an excellent trade-off between accuracy and inference speed.

To demonstrate the effectiveness of our FAST, we conduct extensive experiments on four challenging benchmarks, including Total-Text \cite{ch2017total}, CTW1500 \cite{liu2019curved}, ICDAR 2015 \cite{karatzas2015icdar}, and MSRA-TD500 \cite{yao2012detecting}. 
According to the model size, we name our text detectors FAST-Tiny/Small/Base (short for FAST-T/S/B), respectively.
As shown in Fig.~\ref{fig:f_vs_speed}, on the Total-Text dataset,
FAST-T-448, which means scaling the shorter side of input images to 448 pixels, achieves 81.6\% F-measure at 152.8 FPS, being 1.7\% F-measure higher and 70 FPS faster than the previous fastest method PAN-320~\cite{wang2019efficient}. 
Besides, our best model FAST-B-800 achieves 87.5\% F-measure while still keeping a real-time speed (46.0 FPS).

In summary, our contributions are as follows:

(1) We develop an accurate and efficient arbitrarily-shaped text detector, termed FAST, which is completely GPU-parallel, in terms of post-processing and network architecture.

(2) We propose a minimalist kernel representation (MKR) with a GPU-parallel post-processing---text dilation, significantly reducing its time overhead.

(3) We design a NAS search space and reward function specifically for text detection, and search for a series of backbone networks 
(\ie, TextNet) friendly to text detection with different inference speeds.

(4) Our FAST-T model achieves an astonishing speed of 152.8 FPS while maintaining competitive accuracy on Total-Text. 
With TensorRT~\cite{vanholder2016efficient} optimization, it can be further accelerated to over 600 FPS.

\section{Related Work}

\subsection{Scene Text Detection}
Inspired by general object detection methods~\cite{liu2016ssd,ren2016faster,he2017mask}, many methods~\cite{liao2018textboxes++,liao2017textboxes,liao2018rotation,ma2018arbitrary,shi2017detecting,zhou2017east,tian2016detecting} have been proposed to detect horizontal and multi-oriented text. 
For instance, Tian \etal~\cite{tian2016detecting} presented the CTPN, successfully transferred object detection frameworks for horizontal text detection, and obtained promising results. 
Some researchers~\cite{zhou2017east,liao2018textboxes++,liao2018rotation} have considered the orientations of text lines and designed various methods to detect multi-oriented text. 
However, most of them fail to locate curved text accurately.

To remedy this defect, recent methods cast the text detection task as a segmentation problem.
For example, TextSnake~\cite{long2018textsnake} designed a flexible representation for scene text, which described a text instance as a sequence of ordered and overlapping disks centered at symmetric axes.
PixelLink~\cite{deng2018pixellink} separated adjacent text lines by performing text/non-text prediction and link prediction at the pixel level.
SPCNet~\cite{xie2019scene} and Mask TextSpotter~\cite{lyu2018mask} are designed to detect arbitrarily-shaped text in an instance segmentation manner.
SAE~\cite{tian2019learning} introduced a shape-aware loss and new cluster post-processing to distinguish adjacent text lines with various aspect ratios and small gaps.
PSENet \cite{wang2019shape} proposed to present text instances via text kernels, and developed the progressive scale expansion (PSE) algorithm to merge multi-scale text kernels.
Although the above methods achieve excellent performance, most of them run at a slow inference speed due to the cumbersome post-processing approaches and complicated network architectures.

\begin{figure*}[tbp]
    \centering
    \includegraphics[width=1\linewidth]{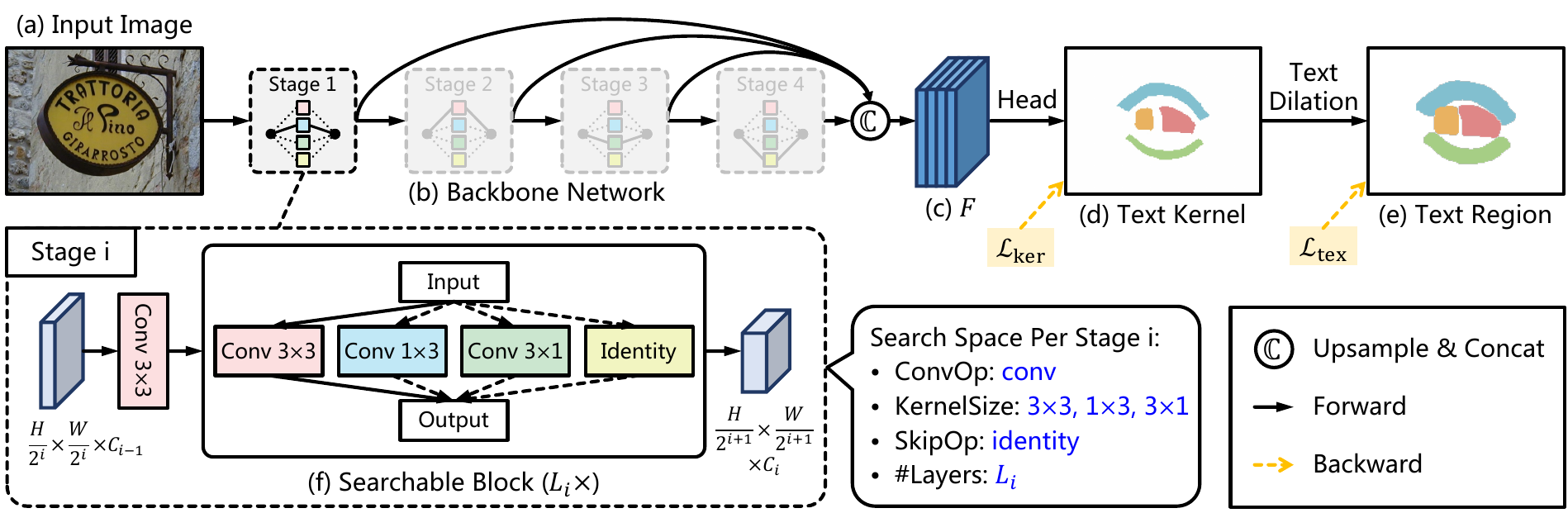}
    \caption{
        Overall architecture of FAST.
        The backbone network is divided into four stages, each of which contains $L_i$ searchable blocks for the architecture search of text detection.
        The multi-scale features from the backbone are upsampled and concatenated as the final feature map $F$, which is used to predict text kernels via a lightweight head~\cite{wang2019efficient} of 2-layer convolutions.
        The GPU-parallel post-processing---text dilation, is applied to reconstruct complete text lines.
        }
    \label{fig:overall_architecture}
\end{figure*}

\subsection{Real-time Text Detection}
With the growing demand of real-time applications, efficient text detection attracts increasing attention.
EAST~\cite{zhou2017east} applied a fully convolutional network (FCN) to directly produce rotated rectangles or quadrangles for text regions, which is the first text detector that runs at 20 FPS.
PAN~\cite{wang2019efficient} and DB~\cite{liao2020real} are two representative real-time text detectors, both of which adopted a lightweight backbone (\ie, ResNet18 \cite{he2016deep}) to speed up inference. 
For post-processing, PAN developed a learnable post-processing algorithm, namely pixel aggregation (PA), to improve the accuracy by using the predicted similarity vectors.
DB proposed the box formation process, which utilized the Vatti clipping algorithm~\cite{vatti1992generic} to dilate the predicted text kernels.
Recently, PAN++~\cite{wang2021pan++} and DB++~\cite{liao2022real} extended their previous methods~\cite{wang2019efficient,liao2020real} and obtained improved detection performance.
Although these methods have simplified the text detection pipeline compared to previous methods~\cite{long2018textsnake,wang2019shape,zhou2017east,xie2019scene}, 
real-time text detection is still room for improvement, due to CPU-based post-processing and sub-optimal hand-crafted network architecture.

\subsection{Neural Architecture Design}
Network architecture design is an ongoing research topic in the field of computer vision~\cite{wang2021pvtv2,ding2021repvgg,chen2022vision}. 
For the text detection task, most of the existing methods~\cite{zhou2017east,wang2019efficient,liao2020real,wang2019shape,tian2016detecting} adopted the hand-crafted backbone networks, such as VGG~\cite{simonyan2014very} and ResNet~\cite{he2016deep}, but these backbones are originally designed for image classification and may not be the best choice for text detection.
Recently, owing to the neural architecture search (NAS) techniques, there has been a significant change in designing neural networks.
Many auto-searched efficient networks, such as Proxyless~\cite{cai2018proxylessnas}, EfficientNet~\cite{tan2019efficientnet}, OFA~\cite{cai2019once}, and MobileNetV3~\cite{howard2019searching}, play increasingly important roles in industry and the research community. 
Despite these developments, these NAS-based models are mainly limited to a few tasks, such as image classification and general object detection, leading to weak generalization ability in other tasks.
To compensate for these drawbacks, many researchers explore applying NAS methods to their specific fields, including semantic segmentation~\cite{liu2019auto}, pose estimation~\cite{xu2021vipnas}, and scene text recognition~\cite{hong2020memory,zhang2020autostr}, \etc.
However, there is still rare to extend NAS approaches to text detection.

\section{Proposed Method}

\begin{figure}[tbp]
    \centering
    \includegraphics[width=0.99\linewidth]{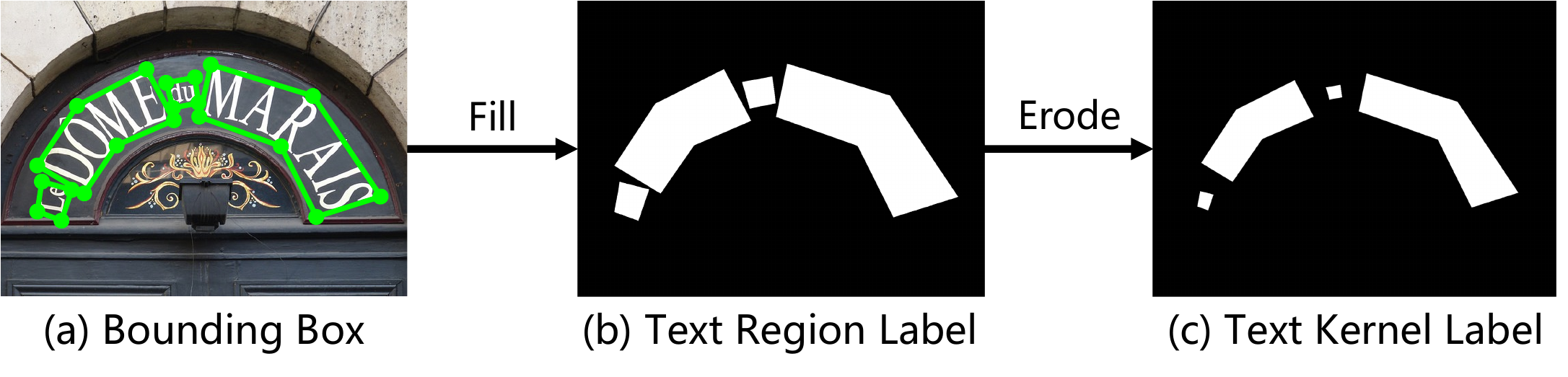}
    \caption{
    Label generation of the minimalist kernel representation (MKR).
    For a given image, the text region label can be generated by filling the bounding boxes. Then,  applying an erosion operator to it, we can obtain the text kernel label.
    We use both these labels to supervise our text detectors.
    }
    \label{fig:label_generation}
\end{figure}

\subsection{Overall Architecture}

As illustrated in Fig. \ref{fig:overall_architecture}, the proposed FAST contains (1) a GPU-parallel post-processing---text dilation, to rebuild complete text lines from predicted text kernels; and (2) a backbone network with multiple searchable blocks  for architecture search of text detection.

In the inference phase, we first feed the input image of $H\!\times\!W\!\times\!3$ into the backbone, and obtain multi-scale features, which are 1/4, 1/8, 1/16, 1/32 of the original image resolution.
Then, we reduce the dimension of each feature map to 128 via 3$\times$3 convolution, and these feature maps are upsampled and concatenated via the function $\mathbb{C(\cdot)}$, to obtain the final feature map $F$, whose shape is $H/4\!\times\!W/4\!\times\!512$ (see Fig. \ref{fig:overall_architecture}(c)).
After that, the final feature map $F$ passes through a lightweight head~\cite{wang2019efficient} of 2-layer convolutions to perform text kernel segmentation.
Finally, we rebuild the complete text regions via the text dilation process with a negligible time overhead, as shown in Fig. \ref{fig:overall_architecture}(d) and Fig. \ref{fig:overall_architecture}(e).

During training, we use loss functions $\mathcal{L}_{\rm ker}$ and $\mathcal{L}_{\rm tex}$ to optimize the text kernel predicted by the network (see Fig. \ref{fig:overall_architecture}(d)) and the text region generated by post-processing (see Fig. \ref{fig:overall_architecture}(e)), respectively.
During searching, we perform architecture search for text detection based on 
the widely-used search framework ProxylessNAS~\cite{cai2018proxylessnas}.
Specifically, we calculate rewards according to the segmentation accuracy and inference speed, and then use the reinforce-based strategy to optimize the network architecture.
The processes of training and searching are performed in an alternative manner. When the  architecture search is finished, we can prune redundant paths and obtain the final architecture.

\subsection{Minimalist Kernel Representation}

\begin{algorithm}[!t]

\caption{PyTorch-like Pseudo Code of Text Dilation}
\label{alg:pseudo_code}
\definecolor{codeblue}{rgb}{0.25,0.5,0.5}
\definecolor{codekw}{rgb}{0.85, 0.18, 0.50}
\lstset{
	backgroundcolor=\color{white},
	basicstyle=\fontsize{8pt}{8pt}\ttfamily\selectfont,
	columns=fullflexible,
	breaklines=true,
	captionpos=b,
	commentstyle=\fontsize{8pt}{8pt}\color{codeblue},
	keywordstyle=\fontsize{8pt}{8pt}\color{codekw},
}

\begin{lstlisting}[language=python]
# s: dilation size

# text dilation for post-processing
def text_dilation(text_kernel, s):

    if not training: # in the inference phase
        # binarize text kernel
        text_kernel = text_kernel > 0 
        # distinguish text kernels using the connected components labeling (CCL) algorithm
        text_kernel = ccl_gpu(text_kernel)
    
    # implement dilation operation with F.max_pool2d
    # args: input, kernel size, stride, padding
    text = F.max_pool2d(text_kernel, s, 1, s//2)    
    
    return text
\end{lstlisting}
\end{algorithm}

\subsubsection{Definition}
To simplify the post-processing, we propose a novel text representation approach termed minimalist kernel representation (MKR).
As illustrated in Fig.~\ref{fig:label_generation}, our MKR formulates a given text line as an eroded text region (\ie, text kernel) with peripheral pixels.
Compared to the existing kernel representations \cite{liao2020real,wang2019shape,wang2019efficient}, our MKR has two main differences as follows.

Firstly, because our text kernel label is generated by the morphological erosion operation, it can be approximatively restored to the complete text region by the reverse operation (\ie, dilation). 
Moreover, both erosion and dilation can be easily implemented in PyTorch with GPU acceleration.

Secondly, our MKR only requires the network to predict a 1-channel output, which is simpler than previous methods that need multi-channel output~\cite{liao2020real,wang2019shape,wang2019efficient}, as illustrated in Fig. \ref{fig:pipeline}.
To our best knowledge, it may be the simplest kernel representation for arbitrarily-shaped text detection.

\subsubsection{Label Generation}
To learn this representation, we need to generate labels for text kernels and text regions.
Specifically, for a given text image, the label of text regions can be directly produced by filling the bounding boxes, which is denoted as $G_{\rm tex}$ (see Fig.~\ref{fig:label_generation}(b)). 
Note that $G_{\rm tex}$ is a binary image, 
applying an erosion operator with $s\times s$ kernel to $G_{\rm tex}$, the peripheral pixels of text regions will be converted to non-text pixels.
To avoid losing text instances due to the erosion operation, we keep at least a minimal text kernel for each text region.
We take this result as the label for text kernels and denote it as $G_{\rm ker}$ (see Fig.~\ref{fig:label_generation}(c)). 

\begin{figure*}[t]
    \centering
\includegraphics[width=1\linewidth]{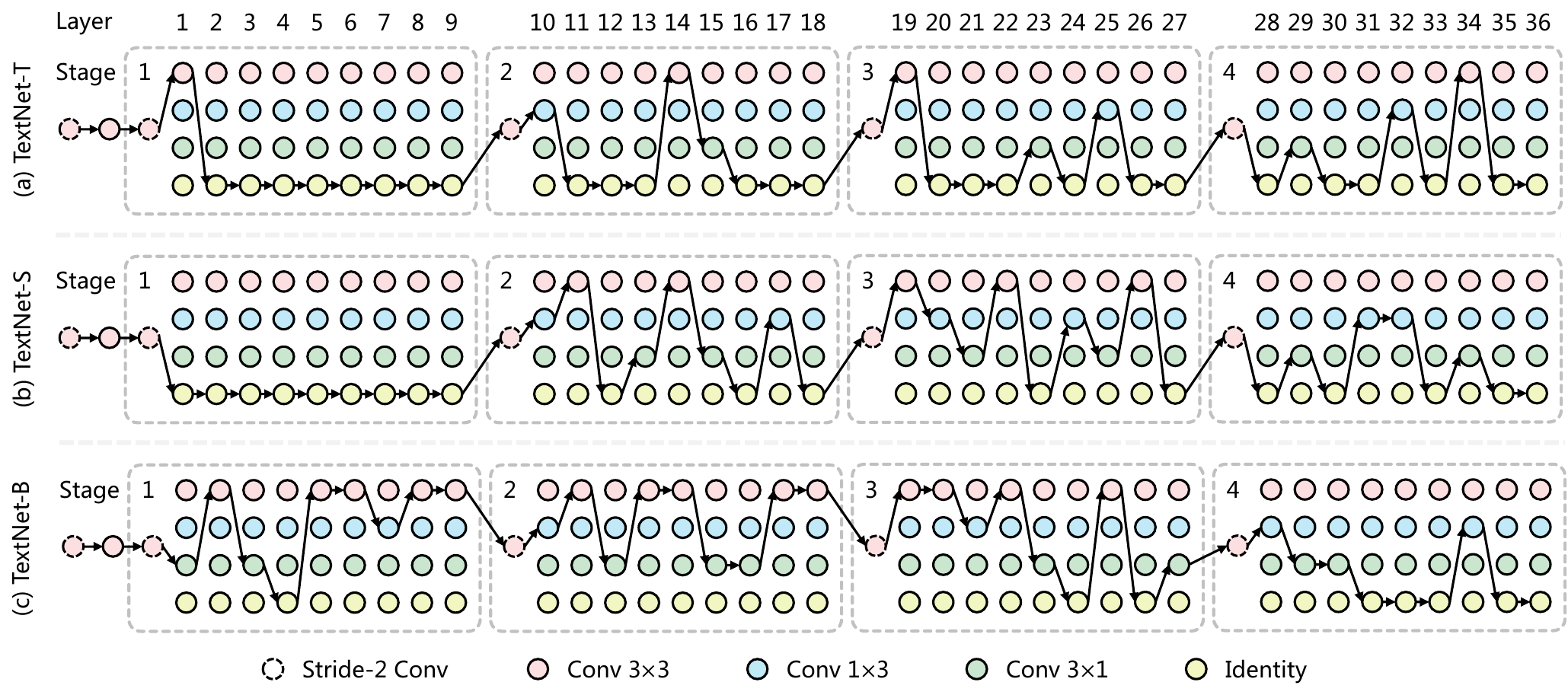}
    \caption{
    Searched backbone architecture of TextNet in our proposed FAST.
    The four nodes in each column represent a searchable block, and the black arrows indicate the selected operations.
    After the architecture search, we can prune redundant paths and obtain the final architecture.
    }
\label{fig:nas_result}
\end{figure*}

\subsubsection{Post-Processing}
Based on the proposed MKR, we develop a GPU-parallel post-processing, termed text dilation, to recover complete text lines with a negligible time overhead. The pseudo code is shown in Algorithm~\ref{alg:pseudo_code}, in which we utilize the max-pooling function with $s\times s$ kernel to implement the dilation operator equivalently.

During training, for a given prediction of text kernels, we directly apply the dilation operator to rebuild whole text regions.
Since this step is differentiable, we can supervise both text kernels and text regions for more accurate predictions, as shown in Fig.~\ref{fig:overall_architecture}.
In the inference phase, we first binarize the predicted text kernels, and implement a GPU-accelerated Connected Components Labeling (CCL) algorithm \cite{allegretti2019optimized} to distinguish different text kernels.
Finally, we apply the dilation operator to reconstruct the complete text lines.

\subsection{Efficient TextNet for Text Detection}

\subsubsection{Search Space}  
Following ProxylessNAS~\cite{cai2018proxylessnas}, we build a backbone network for the architecture search of text detection.
As shown in Fig. \ref{fig:overall_architecture}(f), each stage of the backbone network is comprised of a stride-2 convolution and $L_i$ searchable blocks, where the 3$\times$3 convolution with stride 2 is used to downsample the feature maps, and each searchable block consists of a set of candidate operations, from which the most appropriate one is selected as the final operation after architecture search.
In pursuit of extreme speed, we use reparameterable convolutions~\cite{ding2021repvgg} as the candidate operations, and merge them into plain convolutions without multi-branch topology during inference.

Specifically, we present a layer-level candidate set, defined as \{{\ttfamily conv3$\times$3}, {\ttfamily conv1$\times$3}, {\ttfamily conv3$\times$1}, {\ttfamily identity}\}.
As the 1$\times$3 and 3$\times$1 convolutions have asymmetric kernels and oriented structure priors, they may help to capture the features of extreme aspect-ratio and rotated text lines.
In addition, the identity operator indicates that a layer is skipped, which is used to control the depth and inference speed of the network.
In summary, because there are a total of $L=L_1+L_2+L_3+L_4$ searchable blocks, and each of them has four candidates, the size of the search space is $4^L$.

\subsubsection{Reward Function}
In addition to the search space, we design a customized reward function $\mathcal{R}(\cdot)$, to search network architectures for real-time text detection.
Specifically, given a model $m$, we define the reward function as:
\begin{equation}
	\mathcal{R}(m)=\left( {\rm IoU}_{\rm ker}(m)+\alpha {\rm IoU}_{\rm tex}(m)\right)\times \left(\frac{{\rm FPS}(m)}{T}\right)^w,
	\label{reward_function}
\end{equation}
where ${\rm IoU}_{\rm ker}(m)$ and ${\rm IoU}_{\rm tex}(m)$ denote the intersection-over-union (IoU) metric of the predicted text kernels and text regions, respectively. 
$\alpha$ is the coefficient of ${\rm IoU}_{\rm tex}(m)$, which is empirically set to 0.5.
Besides that, ${\rm FPS}(m)$ means the inference speed of the entire text detector measured on the GPU with batch size 1, and $T$ is the target inference speed. $w$ is a hyper-parameter to balance the accuracy and inference speed, which is set to 0.1 following common practice~\cite{cai2018proxylessnas}.

\subsubsection{Discussion}
Our method is different from existing works on NAS in three main aspects:

$\bullet$  We introduce reparameterable asymmetric convolution~\cite{szegedy2016rethinking,ding2021repvgg} into the search space, which has oriented structure priors that may help to capture the features of extreme aspect-ratio and rotated text lines. 
While most existing NAS methods adopt MBConv~\cite{sandler2018mobilenetv2} as the block, which ignores the geometric characteristics of text lines, and is not efficient enough for GPU.

$\bullet$  We propose a specialized reward function, which considers both the performance of the text kernel and text region, achieving effective architecture search for text detection. 
While most previous reward functions~\cite{tan2019mnasnet,tan2019efficientnet,howard2019searching,wang2020fcos} are designed for image classification or general object detection, and are not suitable for arbitrarily-shaped text detection.

$\bullet$  In this work, we extend the search framework
ProxylessNAS~\cite{cai2018proxylessnas} for text detection. 
We aim to design a faster real-time text detector by compressing the time cost of all components in the text detection pipeline, rather than developing a new search algorithm.
We consider that whether the search algorithm needs to be redesigned for text detection, is an interesting topic that can be further explored in the future.

\begin{table*}[!t]
    \small
    \centering
    \caption{
    Ablation studies of each proposed component in our FAST.
    The shorter sides of images in Total-Text~\cite{ch2017total} and ICDAR 2015~\cite{karatzas2015icdar} are set to 640 and 736 pixels, respectively.}
    \setlength{\tabcolsep}{2.7mm}{
        \begin{tabular}{ll|ll|cc|cc}
        \toprule
        \multirow{2}{*}{Method} & \multirow{2}{*}{\#Param} & \multirow{2}{*}{Backbone} & \multirow{2}{*}{Post-Processing} & \multicolumn{2}{c|}{Total-Text} & \multicolumn{2}{c}{ICDAR 2015} \\
         &  &  &  & F-measure & FPS & F-measure & FPS \\
        \midrule  
        Baseline & 13.0M &ResNet18~\cite{he2016deep} & Pixel Aggregation~\cite{wang2019efficient} & 85.2 & 49.0 & 83.0 & 34.5 \\
        FAST-R18 (ours) & 13.0M & ResNet18~\cite{he2016deep} & Text Dilation & 85.0 \red{(-0.2)} & 62.9 \green{(+13.9)}  & 82.8 \red{(-0.2)}  & 41.4 \green{(+6.9)}\\
        FAST-B (ours) & 10.6M & TextNet-B & Text Dilation & \textbf{86.4 \green{(+1.2)}}  & \textbf{67.5 \green{(+18.5)}} & \textbf{84.7 \green{(+1.7)}} & \textbf{42.7 \green{(+8.2)}} \\
        \bottomrule
        \end{tabular}}
    \label{tab:component}
\end{table*}

\subsection{Loss Function}

The loss function of our FAST can be formulated as:
\begin{equation}
	\mathcal{L}=\mathcal{L}_{\rm ker} + \alpha\mathcal{L}_{\rm tex},
	\label{total_loss}
\end{equation}
where $\mathcal{L}_{\rm ker}$ and $\mathcal{L}_{\rm tex}$ are losses for text kernels and text regions.
Following common practices~\cite{wang2019shape,wang2019efficient},
we apply Dice loss~\cite{milletari2016fully} to supervise the network. Therefore, $\mathcal{L}_{\rm ker}$ and $\mathcal{L}_{\rm tex}$ can be expressed as follows:
\begin{equation}
	\mathcal{L}_{\rm ker}=1-\frac{2\sum_{x,y}P_{\rm ker}(x,y)~G_{\rm ker}(x,y)}{\sum_{x,y}P_{\rm ker}(x,y)^2+\sum_{x,y}G_{\rm ker}(x,y)^2},
	\label{kernel_dice_loss}
\end{equation}

\begin{equation}
	\mathcal{L}_{\rm tex}=1-\frac{2\sum_{x,y}P_{\rm tex}(x,y)~G_{\rm tex}(x,y)}{\sum_{x,y}P_{\rm tex}(x,y)^2+\sum_{x,y}G_{\rm tex}(x,y)^2},
	\label{text_dice_loss}
\end{equation}
where $P(x,y)$ and $G(x,y)$ represent the value of position $(x,y)$ in the prediction and the ground-truth, respectively. 
In addition, we apply Online Hard Example Mining (OHEM) \cite{shrivastava2016training} to $\mathcal{L}_{\rm tex}$ to ignore simple non-text regions.
$\alpha$ balances the importance of $\mathcal{L}_{\rm ker}$ and $\mathcal{L}_{\rm tex}$, which is set to 0.5 in our experiments.

\section{Experiments}

\subsection{Datasets}

\noindent \textbf{Total-Text}~\cite{ch2017total} is a challenging dataset for arbitrarily-shaped text detection, including horizontal, multi-oriented, and curved text lines. 
It contains 1,255 training and 300 testing images, all of which are labeled with polygons at the word level.

\noindent \textbf{CTW1500}~\cite{liu2019curved} is also a widely used dataset for arbitrarily-shaped text detection.
It consists of 1,000 training images and 500 testing images. 
In this dataset, text lines are labeled with 14 points as polygons.

\noindent \textbf{ICDAR 2015}~\cite{karatzas2015icdar} is one of the challenges of the ICDAR 2015 Robust Reading Competition. 
It focuses on multi-oriented text in natural scenes and contains 1,000 training images and 500 testing images. 
The text lines are labeled by quadrangles at the word level.

\noindent \textbf{MSRA-TD500}~\cite{yao2012detecting} is a multi-lingual dataset that contains multi-oriented and long text lines. It has 300 training images and 200 testing images. 
Following the previous works~\cite{long2018textsnake,lyu2018multi,zhou2017east}, we include the 400 images of HUST-TR400~\cite{yao2014unified} as training data.

\noindent \textbf{IC17-MLT}~\cite{nayef2017icdar2017} is a multi-language dataset that consists of 7,200 training images, 1,800 validation images, and 9,000 testing images. In this dataset, text lines 
are annotated with word-level quadrangles.

\subsection{Implementation Details}

\subsubsection{Training Settings} 
Following previous methods \cite{feng2019textdragon,wang2019shape,xie2021polarmask++,xie2019scene}, we pre-train our models on IC17-MLT for 300 epochs, in which images are cropped and resized to 640 $\times$ 640 pixels. 
We then finetune the models for 300 epochs on Total-Text, and 600 epochs on the other three datasets.
The dilation size $s$ is set to 9 when the shorter side is 640 pixels in our experiments.
All models are optimized by Adam~\cite{kingma2014adam} optimizer with batch size 16 on 4 1080Ti GPUs.
We adopt a ``poly" learning rate schedule with an initial learning rate of $1 \times 10^{-3}$.
Training data augmentations include random scale, random crop, random flip, and random rotation.

\subsubsection{Inference Settings}
In the inference phase, we scale the shorter side of images to a fixed size and report the performance on each dataset. 
For a fair comparison, we evaluate all testing images and calculate the average speed.
Our main results are tested with a batch size of 1 on one 1080Ti GPU unless explicitly stated.
When using TensorRT \cite{vanholder2016efficient} to speed up inference, we adopt a V100 GPU instead of 1080Ti to deploy our models, because the 1080Ti GPU does not support half-precision (FP16) inference.

\subsubsection{NAS Settings}
We extend the widely-used ProxylessNAS~\cite{cai2018proxylessnas} for the architecture search of text detection.
During searching, we consider a total of $L = 36$ searchable blocks.
Following the common practice~\cite{he2016deep} of doubling the number of channels when halving the size of feature maps, we set $C_1$, $C_2$, $C_3$, $C_4$ to 64, 128, 256, and 512, respectively. 
We set our target inference speed $T$ as 100, 80, and 60 FPS for searching TextNet-T, -S, and -B, respectively.
To keep generalization ability, we take IC17-MLT~\cite{nayef2017icdar2017} as the training set, and construct a validation set that concludes the training images of ICDAR 2015~\cite{karatzas2015icdar} and Total-Text~\cite{ch2017total}.
The entire network is trained and searched for 200 epochs, which takes around 200 GPU hours on 1080Ti.

\begin{table}[!t]
    \small
    \centering
    \caption{Ablation studies of the erosion/dilation size $s$.
    According to these results, we set the erosion/dilation size $s$ to 9 by default in our experiments.
    }
    \setlength{\tabcolsep}{2.1mm}{
        \begin{tabular}{l|ccccc}
        \toprule
        Dataset & $s=3$ & $s=5$ & $s=7$ & $s=9$ & $s=11$ \\
        \midrule  
        Total-Text~\cite{ch2017total} & 82.3 & 83.4  & 84.8 & \textbf{85.0} & 85.0 \\
        ICDAR 2015~\cite{karatzas2015icdar} & 51.5 & 70.6 & 80.1 & \textbf{82.8} & 82.6 \\
        \bottomrule
        \end{tabular}}
    \label{tab:pooling_size}
\end{table}

\begin{table}[!t]
    \caption{
    Text detection F-measure and inference speed of different backbones on Total-Text~\cite{ch2017total} dataset, where we scale the shorter side of images to 640 pixels.
    Our TextNet models significantly outperform existing hand-crafted and auto-searched networks.
    Note that the \#Param here does not include the classification head.
    }
    \small
    \setlength{\tabcolsep}{1.9mm}
    \begin{tabular}{c|lccccc}
    \toprule
    & Backbone & \#Param & P & R & F & FPS \\
    \cmidrule(r){1-7}
    \multirow{5}{*}{ \rotatebox{90}{Hand-Crafted}}
    & ResNet18~\cite{he2016deep}    & 11.3M & 89.3 & 81.2 & 85.0 & 62.9   \\
    & ResNet50~\cite{he2016deep}    & 23.6M & 89.5 & 81.5 & 85.3 & 32.2   \\
    & ResNet101~\cite{he2016deep}   & 42.6M & \textbf{90.9} & 81.0 & 85.6 & 23.0   \\
    & VGG16~\cite{simonyan2014very} & 14.7M & 88.0 & 81.4 & 84.6 & 25.5   \\
    & PVTv2-B0~\cite{wang2021pyramid} & 3.4M & 90.4 & 79.9 & 84.8 & 24.0   \\
    \cmidrule(r){1-7}
    \multirow{8}{*}{ \rotatebox{90}{Auto-Searched}}
    & EfficientNet-B0~\cite{tan2019efficientnet}    & 3.6M & 90.2 & 81.4 & 85.6 & 39.4  \\
    & Proxyless-GPU~\cite{cai2018proxylessnas}      & 4.6M & 89.6 & 78.1 & 83.5 & 54.7  \\
    & OFANet-12ms~\cite{cai2019once}            & 3.5M & 89.9 & 79.1 & 84.2 & 55.1  \\
    & MobileNetV3~\cite{howard2019searching}  & 3.0M & 90.3 & 78.4 & 83.9 & 56.7  \\
    & GENet-Small~\cite{lin2020neural}  & 6.2M & 89.0 & 77.7  & 83.0  & 74.4 \\
    \cmidrule(r){2-7}
    & TextNet-T (ours) & 6.8M & 87.1 & 81.4 & 84.2 & \textbf{95.5} \\
    & TextNet-S (ours) & 8.0M & 89.1 & 81.9 & 85.4 & 85.3           \\
    & TextNet-B (ours) & 8.9M & 89.9 & \textbf{83.2} & \textbf{86.4} & 67.5            \\
    \bottomrule
    \end{tabular}
    
\label{table:backbone_table}
\end{table}

\subsection{Ablation Study and Analysis} 

\subsubsection{Searched Architecture}
As shown in Fig.~\ref{fig:nas_result}, we plot the searched architectures of TextNet-T/S/B, 
from which we can make the following observations:

$\bullet$   Asymmetric convolutions are the dominant operators in our TextNet, which facilitate the detection of text lines with promising accuracy and high efficiency.

$\bullet$   TextNet-T and -S tend to stack more convolutions in the deep stages (stage-3 and -4), while TextNet-B prefers shallow stages (stage-1 and -2). 
It demonstrates that the stacking rules of TextNet are different under three specified speed constraints (\ie, 100, 80, and 60 FPS),
and shows that the common strategy derived from hand-crafted network design, \ie\ stacking most layers in stage-3~\cite{he2016deep,wang2021pvtv2,wang2022internimage}, may be sub-optimal for real-time text detection.

\subsubsection{Effect of Proposed Components}
We show the effect of each proposed component in Table~\ref{tab:component}. 
For a fair comparison, all models are pre-trained on IC17-MLT~\cite{nayef2017icdar2017} and finetuned on Total-Text~\cite{ch2017total} or ICDAR 2015~\cite{karatzas2015icdar}.
Compared to the baseline that is equipped with ResNet18~\cite{he2016deep} and CPU-based post-processing Pixel Aggregation~\cite{wang2019efficient}, our FAST-R18 replaces the post-processing with the GPU-parallel text dilation, achieving better efficiency with comparable detection performance.  
Moreover, we substitute the ResNet18 backbone with our TextNet-B, which further boosts performance and efficiency, and reduce the number of parameters.

\begin{figure}[t]
    \includegraphics[width=0.98\linewidth]{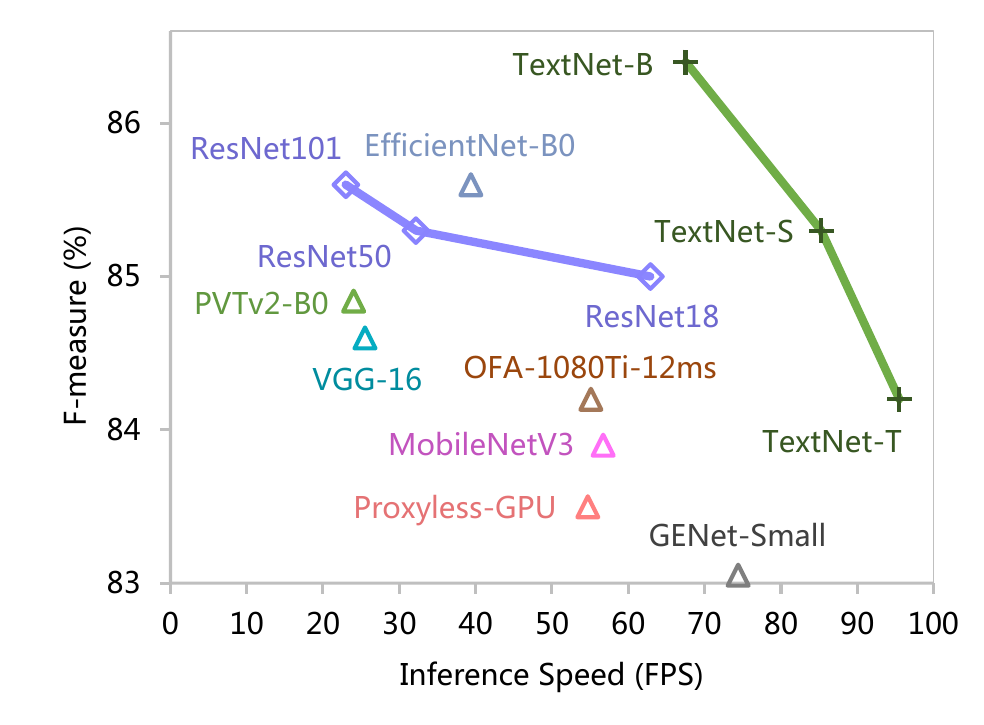}
    \caption{
    Text detection F-measure and inference speed of different backbones on Total-Text~\cite{ch2017total} dataset, where we scale the shorter side of images to 640 pixels.
    Our TextNet models significantly outperform existing hand-crafted and auto-searched networks.
    }
\label{fig:backbone}
\end{figure}

\subsubsection{Effect of Dilation Size}
In this experiment, we study the effect of the dilation size $s$ (equal to the erosion size) based on our FAST-R18 model.
We scale the shorter sides of images in Total-Text~\cite{ch2017total} and ICDAR 2015~\cite{karatzas2015icdar} to 640 and 736 pixels, respectively.
As reported in Table~\ref{tab:pooling_size}, the F-measure on both datasets drops when the dilation size is too small.
Empirically, we set the dilation size $s$ to 9 by default. 
Note that if the size of the shorter side (denoted as $S$) is changed, the dilation size $s$ should be updated proportionally to obtain the best performance:
\begin{equation}
	s_{\rm new}={\rm Round} (S_{\rm new} \times s_{\rm default} / S_{\rm default}).
	\label{new_dilation_size}
\end{equation}
Here, ${\rm Round(\cdot)}$ is a function to round the decimal portion.
For example, when the shorter side is fixed to 800 pixels instead of 640 pixels for training, we set the dilation size $s$ to 11 according to the Eqn.~(\ref{new_dilation_size}).

\subsubsection{Comparison with Hand-Crafted Networks}
We first compare our TextNet with representative hand-crafted backbones, such as ResNets \cite{he2016deep} and VGG16~\cite{simonyan2014very}.
For a fair comparison, all models are first pre-trained on IC17-MLT~\cite{nayef2017icdar2017} and then finetuned on Total-Text~\cite{ch2017total}.
As shown in Fig.~\ref{fig:backbone}, the proposed TextNet models achieve a better trade-off between accuracy and inference speed than previous hand-crafted models by a significant margin.
In addition, notably, our TextNet-T, -S, and -B only have 6.8M, 8.0M, and 8.9M parameters respectively, which are more parameter-efficient than ResNets and VGG16. 
These results demonstrate that TextNet models are effective for text detection on the GPU device. 

\begin{table}[!t]
\small
\centering
\caption{
Detection results on Total-Text~\cite{ch2017total}. 
The suffix of our method means the size of the shorter side.
``Ext." denotes external data.
``P", ``R", and ``F" indicate precision, recall, and F-measure, respectively.
}
\setlength{\tabcolsep}{1.4mm}{

    \begin{tabular}{lcccccc}
    \toprule
    Method & Backbone & Ext.& P & R & F & FPS \\
    \midrule
    \multicolumn{7}{l}{\emph{Non-real-time Methods}} \\
    TextSnake~\cite{long2018textsnake} & VGG16 & \checkmark & 82.7   & 74.5   & 78.4   & 12.4  \\
    TextField~\cite{xu2019textfield}   & VGG16 & \checkmark & 81.2   & 79.9   & 80.6   & -     \\
    PSENet~\cite{wang2019shape}        & ResNet50 & \checkmark & 84.0   & 78.0   & 80.9   & 3.9   \\
    LOMO~\cite{zhang2019look}          & ResNet50 & \checkmark & 88.6   & 75.7   & 81.6   & 4.4   \\
    SPCNet~\cite{xie2019scene}         & ResNet50 & \checkmark & 83.0   & 82.8   & 82.9   & 4.6   \\
    FCENet~\cite{zhu2021fourier}       & ResNet50 & -          & 87.4   & 79.8   & 83.4   & -    \\
    CRAFT~\cite{baek2019character}     & VGG16 & \checkmark & 87.6   & 79.9   & 83.6   & 4.8   \\
    ContourNet~\cite{wang2020contournet} & ResNet50 & -          & 86.9   & 83.9   & 85.4   & 3.8    \\
    TextFuseNet~\cite{ye2020textfusenet} & ResNet101 & \checkmark &  89.0 & 85.3 & 87.1 & 3.3    \\
    ABPNet~\cite{zhang2021adaptive}    & ResNet50 & \checkmark & 90.7   & 85.2   & 87.9   & 10.7    \\
    \midrule
    \multicolumn{7}{l}{\emph{Real-time Methods}} \\
    EAST~\cite{zhou2017east}         & PVANet & -          & 50.0   & 36.2   & 42.0   & -   \\
    PAN~\cite{wang2019efficient}     & ResNet18 & \checkmark & 89.3   & 81.0   & 85.0   & 39.6  \\
    PAN++~\cite{wang2021pan++}       & ResNet18 & \checkmark & 89.9   & 81.0   & 85.3   & 38.3  \\
    DB-R18~\cite{liao2020real}       & ResNet18 & \checkmark & 88.3   & 77.9   & 82.8   & 50.0  \\
    DB-R50~\cite{liao2020real}       & ResNet50 & \checkmark & 87.1   & 82.5   & 84.7   & 32.0  \\
    DB++-R18~\cite{liao2022real}     & ResNet18 & \checkmark & 87.4   & 79.6   & 83.3   & 48.0  \\
    DB++-R50~\cite{liao2022real}     & ResNet50 & \checkmark & 88.9   & 83.2   & 86.0   & 28.0  \\
    \midrule
    FAST-T-448 (ours) & TextNet-T & \checkmark & 86.5 & 77.2 & 81.6 & 152.8 \\
    FAST-S-512 (ours) & TextNet-S & \checkmark & 88.3 & 81.7 & 84.9 & 115.5           \\
    FAST-B-512 (ours) & TextNet-B & \checkmark & 89.6 & 82.4 & 85.8 & 93.2            \\
    FAST-B-640 (ours) & TextNet-B & \checkmark & 89.9 & 83.2 & 86.4 & 67.5            \\
    FAST-B-800 (ours) & TextNet-B & \checkmark & 90.0 & 85.2 & 87.5 & 46.0 \\                
    \bottomrule
    \end{tabular}
    }
    \label{tab:tt}
\end{table}

\subsubsection{Comparison with Auto-Searched Networks}
Here we compare our TextNet with representative auto-searched backbones.
For a fair comparison, all models are pre-trained on IC17-MLT~\cite{nayef2017icdar2017} and finetuned on Total-Text~\cite{ch2017total}.
As shown in Fig. \ref{fig:backbone} and Table~\ref{table:backbone_table}, our TextNet models outperform existing searched networks in terms of accuracy and inference speed, including Proxyless-GPU \cite{cai2018proxylessnas}, OFANet-12ms \cite{cai2019once}, MobileNetV3 \cite{howard2019searching}, GENet-Small~\cite{lin2020neural}, and EfficientNet-B0 \cite{tan2019efficientnet}.
Specifically, TextNet-S achieves 85.4\% F-measure at 85.3 FPS, being 1.2\% more accurate and 1.5$\times$ faster than OFANet-12ms.
TextNet-B yields 86.4\% F-measure at 67.5 FPS, which is 1.7$\times$ faster than EfficientNet-B0 while improving detection performance by 0.8\% F-measure.
A major reason is that these networks are mainly searched for image classification, and the generalization ability on other tasks is not robust.
Therefore, designing the search space and reward function for text detection is meaningful and necessary.

\begin{table}[!t]
    \small
    \centering
    \caption{
    Detection results on CTW1500~\cite{liu2019curved}. 
    The suffix of our method means the size of the shorter side.
    ``Ext." denotes external data.
    ``P", ``R", and ``F" indicate precision, recall, and F-measure, respectively.}
    \setlength{\tabcolsep}{1.4mm}{
    \begin{tabular}{lcccccc}
    \toprule
    Method & Backbone & Ext. & P & R & F & FPS \\
    \midrule
    \multicolumn{7}{l}{\emph{Non-real-time Methods}} \\
    TextSnake~\cite{long2018textsnake}  & VGG16 & \checkmark & 67.9 & 85.3 & 75.6 & - \\
    LOMO~\cite{zhang2019look}           & ResNet50 & \checkmark & 89.2 & 69.6 & 78.4 & 4.4 \\
    SAE~\cite{tian2019learning}         & ResNet50 & \checkmark & 82.7 & 77.8 & 80.1 & - \\
    TextField~\cite{xu2019textfield}    & VGG16 & \checkmark & 83.0 & 79.8 & 81.4 & - \\
    PSENet~\cite{wang2019shape}         & ResNet50 & \checkmark & 84.8 & 79.7 & 82.2 & 3.9 \\
    FCENet~\cite{zhu2021fourier}        & ResNet50 & -          & 85.7 & 80.7 & 83.1 & -    \\
    CRAFT~\cite{baek2019character}      & VGG16 & \checkmark & 86.0 & 81.1 & 83.5 & 7.6 \\
    ContourNet~\cite{wang2020contournet}& ResNet50 & -          & 83.7 & 84.1 & 83.9 & 4.5 \\
    ReLaText~\cite{ma2021relatext}      & ResNet50 & \checkmark & 86.2 & 83.3 & 84.8 & 10.6    \\
    ABPNet~\cite{zhang2021adaptive}     & ResNet50 & \checkmark & 86.5 & 83.6 & 85.0 & 12.2    \\
    TextFuseNet~\cite{ye2020textfusenet} & ResNet101 & \checkmark &  87.8 & 85.4 & 86.6 & 3.7  \\
    \midrule
    \multicolumn{7}{l}{\emph{Real-time Methods}} \\
    EAST~\cite{zhou2017east}           & PVANet & -          & 78.7 & 49.1 & 60.4 & 21.2 \\
    PAN~\cite{wang2019efficient}       & ResNet18 & \checkmark & 86.4 & 81.2 & 83.7 & 39.8 \\
    PAN++~\cite{wang2021pan++}         & ResNet18 & \checkmark & 87.1 & 81.1 & 84.0 & 36.0 \\
    DB-R18~\cite{liao2020real}         & ResNet18     & \checkmark & 84.8 & 77.5 & 81.0 & 55.0 \\
    DB-R50~\cite{liao2020real}         & ResNet50     & \checkmark & 86.9 & 80.2 & 83.4 & 22.0 \\
    DB++-R18~\cite{liao2022real}       & ResNet18     & \checkmark & 86.7 & 81.3 & 83.9 & 40.0 \\
    DB++-R50~\cite{liao2022real}       & ResNet50     & \checkmark & 88.5 & 82.0 & 85.1 & 21.0 \\
    \midrule
    FAST-T-512 (ours) & TextNet-T & \checkmark & 85.5 & 77.9 & 81.5 & 129.1    \\
    FAST-S-512 (ours) & TextNet-S & \checkmark & 85.6 & 78.7 & 82.0 & 112.9    \\
    FAST-B-512 (ours) & TextNet-B & \checkmark & 85.7 & 80.2 & 82.9 & 92.6    \\
    FAST-B-640 (ours) & TextNet-B & \checkmark & 87.8 & 80.9 & 84.2 & 66.5    \\    
    \bottomrule
    \end{tabular}}
\label{tab:ctw}
\end{table}

\subsection{Comparison with State-of-the-Art Methods}

\subsubsection{Curve Text Detection}
To show the advantages of FAST in detecting curved text, we compare it with existing state-of-the-art methods on the Total-Text \cite{ch2017total} and CTW1500 \cite{yuliang2017detecting} datasets, and report the results in Table~\ref{tab:tt} and Table~\ref{tab:ctw}.

On Total-Text, FAST-T-448 yields an F-measure of 81.6\% at 152.8 FPS, which is faster than all previous methods.
Our FAST-S-512 outperforms the real-time text detector DB++-R18 \cite{liao2022real} by 1.6\% in F-measure (84.9\% \emph{vs.} 83.3\%) and runs 2.4$\times$ faster.
Compared to PAN++ \cite{wang2021pan++}, FAST-B-640 is 29.2 FPS faster, while the F-measure is 1.1\% better (86.4\% \emph{vs.} 85.3\%). 
It is notable that when taking a larger input resolution, FAST-B-800 achieves the best F-measure of 87.5\%, surpassing all real-time counterparts in F-measure by at least 1.5\% while still keeping a fast inference speed (46.0 FPS).

Similar results are also on CTW1500.
For example, the inference speed of FAST-T-512 is 129.1 FPS, which is at least 2.3$\times$ faster than prior arts, while the F-measure is still very competitive (81.5\%). 
The best F-measure of our method is 84.2\%, which is slightly higher than the strong counterpart DB++-R18~\cite{liao2022real} (84.2\% \emph{vs.} 83.9\%), while our method runs at a faster speed (66.5 FPS \emph{vs.} 40.0 FPS).
We show some qualitative curved text detection results in Fig.~\ref{fig:visual}(a)(b), which demonstrates that the proposed FAST can accurately locate text lines with complex shapes.

\begin{table}[!t]
    \caption{
Detection results on ICDAR 2015~\cite{karatzas2015icdar}.
The suffix of our method means the size of the shorter side.
``Ext." denotes external data.
``P", ``R", and ``F" indicate precision, recall, and F-measure, respectively.}
    \small
    \setlength{\tabcolsep}{1.5mm}{
    \begin{tabular}{lcccccc}
        \toprule
        Method & Backbone & Ext. & P & R & F & FPS \\
        \midrule
        \multicolumn{7}{l}{\emph{Non-real-time Methods}} \\
        Corner~\cite{lyu2018multi}           & VGG16 & \checkmark & 94.1 & 70.7 & 80.7 & 3.6   \\
        PixelLink~\cite{deng2018pixellink}   & VGG16 & -          & 82.9 & 81.7 & 82.3 & 7.3   \\
        TextSnake~\cite{long2018textsnake}   & VGG16 & \checkmark & 84.9 & 80.4 & 82.6 & 1.1   \\
        FCENet~\cite{zhu2021fourier}         & ResNet50 & -          & 85.1 & 84.2 & 84.6 & -    \\
        PolarMask++~\cite{xie2021polarmask++}& ResNet50 & \checkmark & 87.3 & 83.5 & 85.4 & 10.0 \\
        PSENet~\cite{wang2019shape}          & ResNet50 & \checkmark & 86.9 & 84.5 & 85.7 & 1.6   \\
        CRAFT~\cite{baek2019character}       & VGG16 & \checkmark & 89.8 & 84.3 & 86.9 & -    \\
        LOMO~\cite{zhang2019look}            & ResNet50 & \checkmark & 91.3 & 83.5 & 87.2 & 3.4   \\
        SPCNet~\cite{xie2019scene}           & ResNet50 & \checkmark & 88.7 & 85.8 & 87.2 & 4.6   \\
        \midrule
        \multicolumn{7}{l}{\emph{Real-time Methods}} \\
        EAST~\cite{zhou2017east}        & PVANet  & -          & 83.6 & 73.5 & 78.2 & 13.2  \\
        PAN~\cite{wang2019efficient}    & ResNet18  & \checkmark & 84.0 & 81.9 & 82.9 & 26.1  \\
        PAN++~\cite{wang2021pan++}      & ResNet18  & \checkmark & 85.9 & 80.4 & 83.1 & 28.2  \\
        DB-R18~\cite{liao2020real}      & ResNet18  & \checkmark & 86.8 & 78.4 & 82.3 & 48.0  \\
        DB-R50~\cite{liao2020real}      & ResNet50  & \checkmark &  91.8 & 83.2 & 87.3 & 12.0  \\
        DB++-R18~\cite{liao2022real}    & ResNet18  & \checkmark & 90.1 & 77.2 & 83.1 & 44.0  \\
        DB++-R50~\cite{liao2022real}    & ResNet50  & \checkmark & 90.9 & 83.9 & 87.3 & 10.0  \\
        \midrule
        FAST-T-736 (ours) & TextNet-T & \checkmark & 86.0 & 77.9 & 81.7 & 60.9  \\
        FAST-S-736 (ours) & TextNet-S & \checkmark & 86.3 & 79.8 & 82.9 & 53.9  \\
        FAST-B-736 (ours) & TextNet-B & \checkmark & 88.0 & 81.7 & 84.7 & 42.7  \\
        FAST-B-896 (ours) & TextNet-B & \checkmark & 89.2 & 83.6 & 86.3 & 31.8  \\
        FAST-B-1280 (ours) & TextNet-B & \checkmark & 89.7 & 84.6 & 87.1 & 15.7  \\                
        \bottomrule
    \end{tabular}}
\label{tab:ic15}
\end{table}

\subsubsection{Oriented Text Detection}
We evaluate the effectiveness of FAST in detecting oriented text lines on the ICDAR 2015 \cite{karatzas2015icdar} dataset.
From Table~\ref{tab:ic15}, we can observe that our fastest model FAST-T-736 reaches 60.9 FPS and maintains a competitive F-measure of 81.7\%. 
Compared with PAN++~\cite{wang2021pan++}, FAST-B-896 surpasses it by 3.2\% in F-measure (86.3\% \emph{vs.} 83.1\%) and is more efficient (31.8 FPS \emph{vs.} 28.2 FPS). 
Because ICDAR 2015 contains many small text lines, previous
methods~\cite{xie2019scene,wang2019shape} always adopt high-resolution images to ensure detection performance.
With this setting, FAST-B-1280 achieves an F-measure of 87.1\%, which is comparable with DB-R50~\cite{liao2020real} and DB++-R50~\cite{liao2022real} (87.1\% \emph{vs.} 87.3\%). Besides, compared with PSENet \cite{wang2019shape}, this model outperforms it by 1.4\% F-measure (87.1\% \emph{vs.} 85.7\%) and runs 9.8$\times$ faster. 
Some qualitative results of oriented text detection are shown in Fig.~\ref{fig:visual}(c).

\begin{table}[!t]
    \small
    \caption{
Detection results on MSRA-TD500~\cite{yao2012detecting}.
The suffix of our method means the size of the shorter side.
``Ext." denotes external data.
``P", ``R", and ``F" indicate precision, recall, and F-measure, respectively.}
    \setlength{\tabcolsep}{1.5mm}{
        \begin{tabular}{lcccccc}
        \toprule
        Method & Backbone & Ext. & P & R & F & FPS \\
        \midrule
        \multicolumn{7}{l}{\emph{Non-real-time Methods}} \\
        PixelLink~\cite{deng2018pixellink} & VGG16 & -          & 83.0 & 73.2 & 77.8 & 3.0 \\
        TextSnake~\cite{long2018textsnake} & VGG16 & \checkmark & 83.2 & 73.9 & 78.3 & 1.1 \\
        TextField~\cite{xu2019textfield}   & VGG16 & \checkmark & 87.4 & 75.9 & 81.3 & 5.2 \\
        Corner~\cite{lyu2018multi}         & VGG16 & \checkmark & 87.6 & 76.2 & 81.5 & 5.7 \\
        CRAFT~\cite{baek2019character}     & VGG16 & \checkmark & 88.2 & 78.2 & 82.9 & 8.6 \\
        SAE~\cite{tian2019learning}        & ResNet50 & \checkmark & 84.2 & 81.7 & 82.9 & 3.0    \\
        SBD~\cite{liu2019omnidirectional}  & ResNet50 & \checkmark & 89.6 & 80.5 & 84.8 & 3.2    \\
        DRRG~\cite{zhang2020deep}          & VGG16 & \checkmark & 88.0 & 82.3 & 85.1 & - \\
        ABPNet~\cite{zhang2021adaptive}    & ResNet50 & \checkmark & 86.6 & 84.5 & 85.6 & 12.3    \\
        ReLaText~\cite{ma2021relatext}     & ResNet50 & \checkmark & 90.5 & 83.2 & 86.7 & 8.3    \\
        \midrule
        \multicolumn{7}{l}{\emph{Real-time Methods}} \\
        EAST~\cite{zhou2017east}          & PVANet & -          & 87.3 & 67.4 & 76.1 & 13.2 \\
        PAN~\cite{wang2019efficient}      & ResNet18 & \checkmark & 84.4 & 83.8 & 84.1 & 30.2 \\
        PAN++~\cite{wang2021pan++}        & ResNet18 & \checkmark & 85.3 & 84.0 & 84.7 & 32.5 \\
        DB-R18~\cite{liao2020real}        & ResNet18 & \checkmark & 90.4 & 76.3 & 82.8 & 62.0 \\
        DB-R50~\cite{liao2020real}        & ResNet50 & \checkmark & 91.5 & 79.2 & 84.9 & 32.0 \\
        DB++-R18~\cite{liao2022real}        & ResNet18 & \checkmark & 87.9 & 82.5 & 85.1 & 55.0 \\
        DB++-R50~\cite{liao2022real}        & ResNet50 & \checkmark & 91.5 & 83.3 & 87.2 & 29.0 \\
        \midrule
        FAST-T-512 (ours) & TextNet-T & \checkmark & 91.1 & 78.8 & 84.5 & 137.2    \\
        FAST-T-736 (ours) & TextNet-T & \checkmark & 88.1 & 81.9 & 84.9 & 79.6 \\
        FAST-S-736 (ours) & TextNet-S & \checkmark & 91.6 & 81.7 & 86.4 & 72.0 \\
        FAST-B-736 (ours) & TextNet-B & \checkmark & 92.1 & 83.0 & 87.3 & 56.8 \\
        \bottomrule
        \end{tabular}
    }
\label{tab:msra}
\end{table}

\subsubsection{Long Straight Text Detection}
FAST is also robust for long straight text detection.
As reported in Table~\ref{tab:msra}, on the MSRA-TD500 \cite{yao2012detecting} dataset, 
FAST-T-736 runs at 137.2 FPS with 84.5\% F-measure, which is more efficient than all previous real-time detectors. 
For example, it is 4.5$\times$ and 2.2$\times$ faster than PAN~\cite{wang2019efficient} and DB-R18~\cite{liao2020real} respectively, while keeping higher detection F-measure.
Besides, FAST-S-736 achieves 86.4\% F-measure at 72.0 FPS, outperforming DB++-R18~\cite{liao2022real} by 1.3\% (86.4\% \emph{vs.} 85.1\%) and runs 17 FPS faster (72.0 FPS \emph{vs.} 55.0 FPS).
FAST-B-736 yields the F-measure of 87.3\%, which is slightly better than DB++-R50~\cite{liao2022real} but with significantly higher efficiency (56.8 FPS \emph{vs.} 29.0 FPS).
We present some qualitative straight text detection results in Fig.~\ref{fig:visual}(d).

\subsection{Efficiency Analysis}
\subsubsection{Parameter}
In Table~\ref{tab:more_comparison}, we compare the number of parameters of the proposed FAST with representative arbitrarily-shaped text detectors.
As shown, our detectors FAST-T, FAST-S, and FAST-B have 8.5M, 9.7M, and 10.6M parameters respectively, which are more parameter-efficient than previous real-time text detectors, such as PAN~\cite{wang2019efficient} (12.2M) and DB-R18~\cite{liao2020real} (13.8M).

\subsubsection{FLOPs}
We report the total FLOPs of different text detectors in Table~\ref{tab:more_comparison}. Although our primary concern is inference speed rather than FLOPs, our FAST models also have advantages over previous methods. 
For example, on the Total-Text dataset~\cite{ch2017total}, with similar FLOPs as DB-R18~\cite{liao2020real} (29.5G \emph{vs.} 30.0G), our FAST-T-512 improves the F-measure from 82.8\% to 83.5\%.
Besides,
FAST-B-640 achieves an F-measure of 86.4\% with 64.0G FLOPs, being 1.4 points better (86.4\% \emph{vs.} 85.0\%) than PAN-640~\cite{wang2019efficient} that has similar FLOPs (64.0G \emph{vs.} 65.4G).
Compared with PSENet~\cite{wang2019shape}, our FAST-B-800 achieves the more competitive performance of 87.5\% F-measure, but its computational cost is less than a third of the PSENet (100.1G \emph{vs.} 345.9G).

\subsubsection{Inference Speed}
In this experiment, we adopt TensorRT \cite{vanholder2016efficient}, an inference engine specially designed for industrial deployment, to further speed up our FAST models.
Because the 1080Ti GPU does not support half-precision (FP16) inference, we consider two additional settings: (1) V100 + PyTorch (FP32) with batch size 1, and (2) V100 + TensorRT (FP16) with batch size from 1 to 8. 

As shown in Table~\ref{tab:more_comparison}, when using PyTorch and V100 GPU, our fastest model FAST-T-448 reaches a speed of 187.0 FPS.
With TensorRT optimization, it can be further accelerated to 634.7 FPS (batch size = 8), being 3.4$\times$ faster than its PyTorch implementation (batch size = 1), which demonstrates the efficiency of FAST in practical applications.

\begin{table*}[t]
\small
\centering

\caption{
Efficiency analysis with state-of-the-art methods on Total-Text~\cite{ch2017total}.
``F" indicates F-measure. 
``Scale" denotes the scale of the testing image, where ``$L$:" means the longer side is fixed, ``$S$:" means the shorter side is fixed, and ``$H$:" means the height is fixed.
``$BS$" is short for batch size. 
``\#Param" represents the total number of parameters for the detectors.
The PyTorch FPS is measured with batch size 1.
}
\setlength{\tabcolsep}{2.2mm}{
    \begin{tabular}{lcccccccccc}
    \toprule
    \multirow{2}{*}{Method} & \multirow{2}{*}{F-measure} & \#Param & \multirow{2}{*}{Scale} & \multirow{2}{*}{FLOPs} & \multicolumn{2}{c}{PyTorch FPS (FP32)} &\multicolumn{4}{c}{TensorRT FPS (V100 FP16)} \\
      &  & (total)& &  & 1080Ti & V100 & $\rm BS=1$ & $\rm BS=2$ & $\rm BS=4$ & $\rm BS=8$  \\
    \midrule
    PSENet-4s~\cite{wang2019shape} & 79.6 & 28.7M & $L$: 1280 & 345.9G & 8.4 & 10.8 & 59.3& 69.2 & 72.8 & 77.3 \\ 
    \midrule
    PAN-320~\cite{wang2019efficient} & 79.9 & \multirow{3}{*}{12.2M} &  $S$: 320  & \textbf{16.4G} & 82.4 & 110.8 & 237.5 & 323.2 & 363.4 & 385.2\\
    PAN-512~\cite{wang2019efficient}  & 84.3 &   & $S$: 512 & 41.9G & 57.1 & 85.5 & 187.6 & 204.3 & 215.5 & 243.8\\
    PAN-640~\cite{wang2019efficient}  & 85.0 &  & $S$: 640 & 65.4G & 39.6 & 64.6& 130.4 & 154.4 & 162.8 & 180.7 \\
    \midrule
    DB-R18~\cite{liao2020real} & 82.8 & 13.8M &  $H$: 640  & 30.0G & 50.0 & 73.0 & 105.4 & 143.7 & 204.5 & 227.6\\
    \midrule
    FAST-T-448 (ours) & 81.6 & \multirow{3}{*}{\textbf{8.5M}} & $S$: 448 & 22.6G & \textbf{152.8} & \textbf{187.0}& \textbf{385.9} & \textbf{524.8} & \textbf{628.1} & \textbf{634.7} \\
    FAST-T-512 (ours)  & 83.5 &  & $S$: 512 & 29.5G & 131.1 &176.4 & 343.4 & 408.2 & 498.5 & 525.6 \\
    FAST-T-640 (ours)  & 84.2 &  & $S$: 640 & 46.0G & 95.5 & 136.4 & 275.6 & 287.1 & 345.4 & 364.0    \\
    \midrule
    FAST-S-512 (ours) & 84.9 & \multirow{2}{*}{9.7M} & $S$: 512 & 33.1G & 115.5 & 150.5 &289.7  & 400.2 & 465.7 & 481.5\\
    FAST-S-640 (ours)  & 85.4  &  & $S$: 640 & 51.7G & 85.3 & 115.9 &249.6 & 301.7 & 325.0 & 332.0  \\
    \midrule
    FAST-B-512 (ours) & 85.8 & \multirow{3}{*}{10.6M} & $S$: 512 & 41.0G & 93.2 & 128.6 & 281.0 & 358.6 & 414.3 & 448.0 \\
    FAST-B-640 (ours)  & 86.4 &  & $S$: 640 & 64.0G  & 67.5 &99.8 &219.7 & 261.8 & 293.3 & 297.0 \\
    FAST-B-800 (ours)  & \textbf{87.5} &  & $S$: 800 & 100.1G & 46.0 &69.2 & 143.0 & 177.5 & 189.2 & 195.3 \\
    \bottomrule
    \end{tabular}}
    \label{tab:more_comparison}
\end{table*}

\begin{figure*}[!t]
    \centering
    \includegraphics[width=0.9\linewidth]{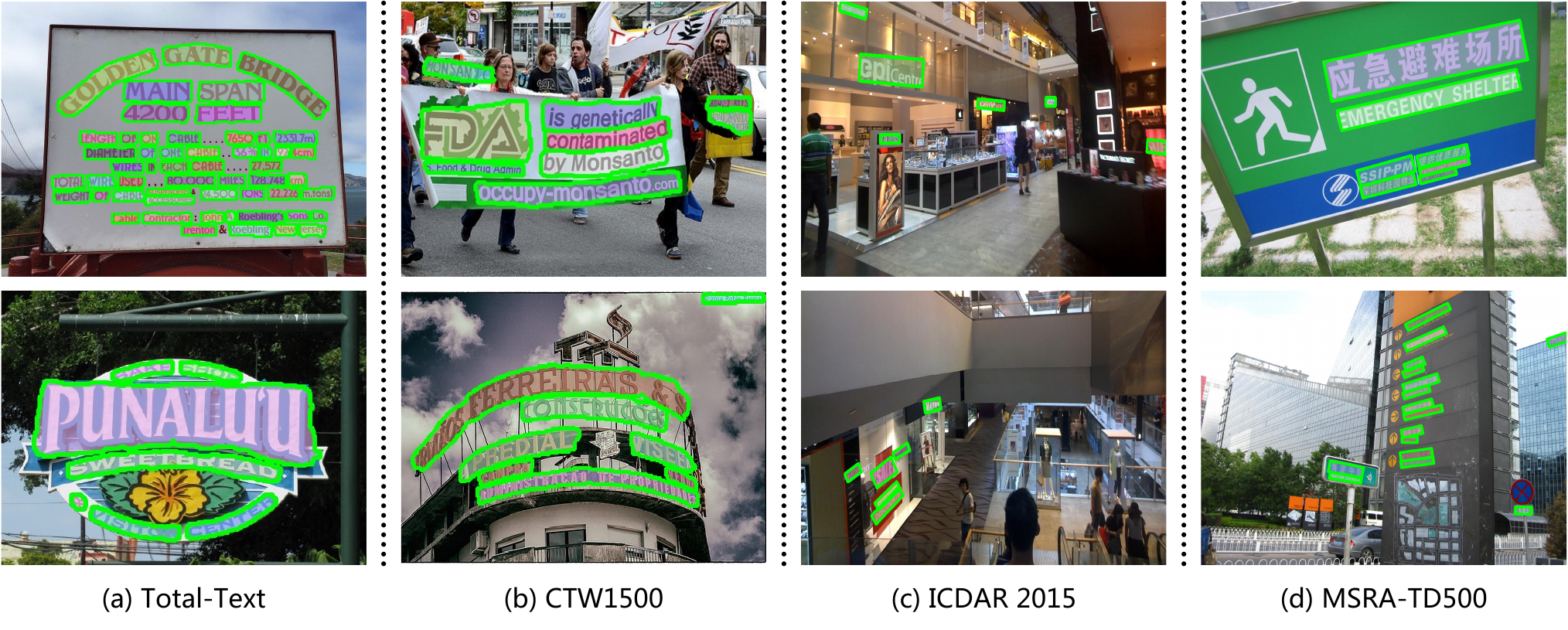}
    \caption{
    Qualitative text detection results of FAST on Total-Text \cite{ch2017total}, CTW1500 \cite{yuliang2017detecting}, ICDAR 2015 \cite{karatzas2015icdar} and MSRA-TD500 \cite{yao2012detecting}.
    These results show that our FAST models are suitable for various complicated natural scenes, including arbitrary shapes, multiple languages, and extreme aspect ratios.
    }
    \label{fig:visual}
\end{figure*}

\subsection{Qualitative Results}

In this section, we show some qualitative text detection results of the proposed FAST on four challenging datasets, including Total-Text (see Fig.~\ref{fig:visual}(a)), CTW1500 (see Fig.~\ref{fig:visual}(b)), ICDAR 2015 (see Fig.~\ref{fig:visual}(c)) and MSRA-TD500 (see Fig.~\ref{fig:visual}(d)). 
These results demonstrate that our FAST models are suitable for various complicated natural scenes, including arbitrary shapes, multiple languages, extreme aspect ratios, \etc.

\section{Conclusion}

In this work, we proposed FAST, a faster arbitrarily-shaped text detector.
To achieve high efficiency,
we presented a minimalist kernel representation (MKR), as well as a GPU-parallel post-processing---text dilation, making our models can completely run on the GPU.
Moreover, we designed a search space and reward function tailored for text detection, and searched for a series of efficient backbone networks (\ie, TextNet) friendly to text detection.
Extensive experiments on several challenging datasets show that equipped with these two designs, our FAST achieves a significantly better trade-off between detection performance and inference speed than previous works. 
We hope our method could serve as a cornerstone for text-related real-time applications.


\newpage

\bibliographystyle{IEEEtran}
\bibliography{egbib}

\newpage

 




\vfill

\end{document}